\documentclass[10pt,twocolumn,letterpaper]{article}

\usepackage[pagenumbers]{cvpr} 

\definecolor{cvprblue}{rgb}{0.21,0.49,0.74}
\usepackage[pagebackref,breaklinks,colorlinks,allcolors=cvprblue]{hyperref}

\usepackage[table]{xcolor}
\usepackage{multirow}
\usepackage{array} 

\def\paperID{10036} 
\def\confName{CVPR}
\def\confYear{2026}

\title{CATS-V2V: A Real-World Vehicle-to-Vehicle Cooperative Perception Dataset with Complex Adverse Traffic Scenarios}

\author{Hangyu Li$^{1,*}$, Bofeng Cao$^{1,}$\thanks{Equal contribution} , Zhaohui Liang$^{1}$, Wuzhen Li$^{1}$, Juyoung Oh$^{1}$, Yuxuan Chen$^{1}$, \\
Shixiao Liang$^{1}$, Hang Zhou$^{1}$, Chengyuan Ma$^{1}$, Jiaxi Liu$^{1}$, Zheng Li$^{1}$, Peng Zhang$^{1}$, KeKe Long$^{1}$, \\
Maolin Liu$^{2}$, Jackson Jiang$^{2}$, Chunlei Yu$^{2}$, Shengxiang Liu$^{2}$, Hongkai Yu$^{3}$, Xiaopeng Li$^{1}$\\
$^1$University of Wisconsin-Madison \quad $^{2}$wuwen-ai \quad $^{3}$Cleveland State University \\
}

\begin{document}

\twocolumn[{%
\renewcommand\twocolumn[1][]{#1}%
\maketitle
\vspace{-2em}
\centering
\includegraphics[width = \linewidth]{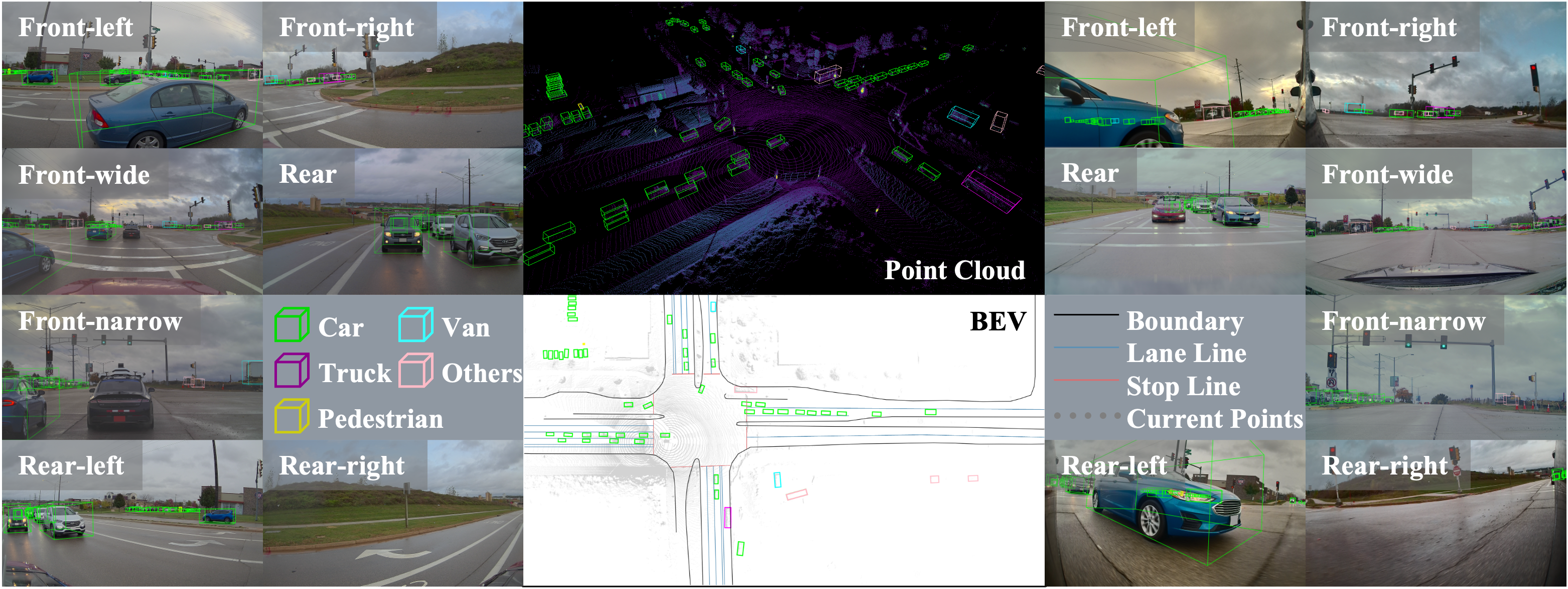}
\captionof{figure}{One frame of CATS-V2V dataset. The middle-upper shows the combined point cloud and 3D bounding box annotations, while the middle-lower presents the HD map and BEV annotations. On sides are the seven camera views of each vehicle with projected annotations.}
\vspace{1em}
\label{fig:combine}
}]

\begin{abstract}
Vehicle-to-Vehicle (V2V) cooperative perception has great potential to enhance autonomous driving performance by overcoming perception limitations in complex adverse traffic scenarios (CATS). Meanwhile, data serves as the fundamental infrastructure for modern autonomous driving AI. However, due to stringent data collection requirements, existing datasets focus primarily on ordinary traffic scenarios, constraining the benefits of cooperative perception. To address this challenge, we introduce CATS-V2V, the first-of-its-kind real-world dataset for V2V cooperative perception under complex adverse traffic scenarios. The dataset was collected by two hardware time-synchronized vehicles, covering 10 weather and lighting conditions across 10 diverse locations. The 100-clip dataset includes 60K frames of 10 Hz LiDAR point clouds and 1.26M multi-view 30 Hz camera images, along with 750K anonymized yet high-precision RTK-fixed GNSS and IMU records. Correspondingly, we provide time-consistent 3D bounding box annotations for objects, as well as static scenes to construct a 4D BEV representation. On this basis, we propose a target-based temporal alignment method, ensuring that all objects are precisely aligned across all sensor modalities. We hope that CATS-V2V, the largest-scale, most supportive, and highest-quality dataset of its kind to date, will benefit the autonomous driving community in related tasks.
\end{abstract}
\section{Introduction}
\label{sec:intro}

\begin{figure*}[htbp]
  \centering
  \includegraphics[width = \linewidth]{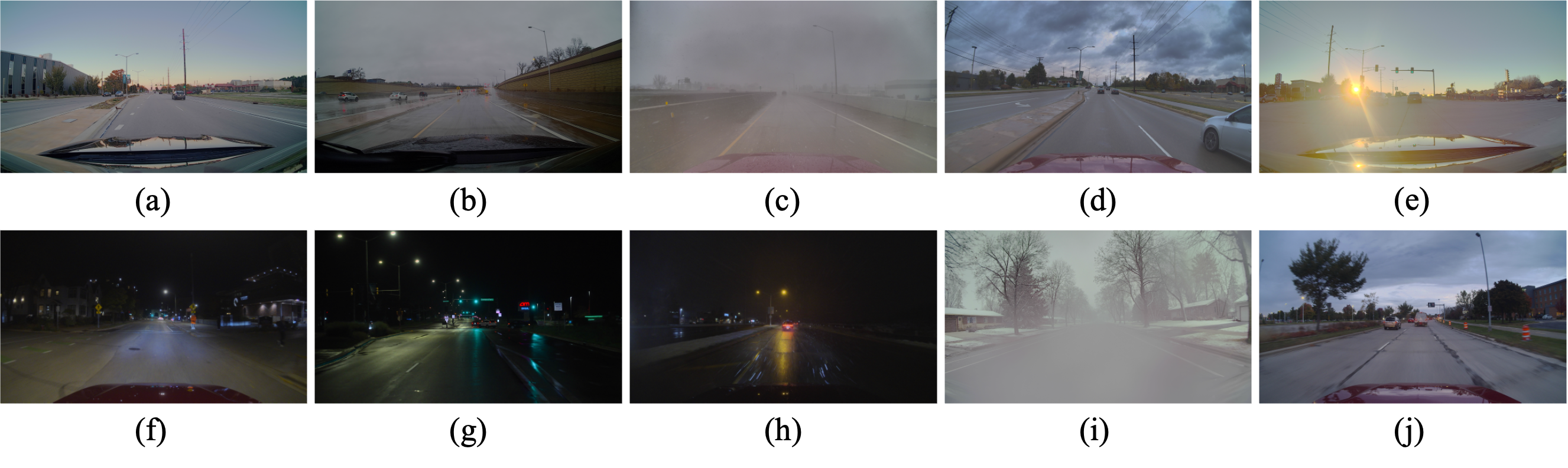}
  \caption{10 CATS scenarios at 10 locations: (a) Clear day at an arterial road; (b) Rainy day at a highway; (c) Snowy day at a highway; (d) Dawn at an arterial road; (e) Dusk with direct sunlight at an arterial road; (f) Clear night at a collector road; (g) Rainy night at ab arterial road; (h) Snwoy night at an arterial road; (i) Foggy day at a local street; (j) Overcast day with work zone at an arterial road.}
\label{fig:scenarios}
\end{figure*}

Cooperative perception has emerged as a promising approach to overcoming the sensing limitations of autonomous vehicles in rare but safety-critical corner cases \cite{han2023collaborative,ma4973353real}. Onboard sensors such as cameras and LiDARs are inherently constrained by occlusion, noise, and limited range, which can severely degrade perception performance in Complex Adverse Traffic Scenarios (CATS) \cite{fan2023autonomous,li2025robotic}.

CATS encompass adverse weather and illumination conditions, irregular work zones, and visually challenging scenes that often cause sensor failures and decision-making instability \cite{yoneda2019automated,zheng2023robust,sezgin2023safe}. These scenarios constitute the long-tail distribution of real-world driving, where the robustness and generalization of autonomous driving systems are most critically tested \cite{feng2021intelligent,feng2023dense}.

To mitigate these issues, cooperative perception (CP) allows vehicles and infrastructure to share perceptual information, extending situational awareness beyond a single vehicle's view \cite{wang2020v2vnet,chen2019f,xu2022v2x}. Depending on the source, it can be categorized into Vehicle-to-Vehicle (V2V), Vehicle-to-Infrastructure (V2I), or hybrid V2X modes. Among them, V2V cooperative perception offers the greatest flexibility and scalability since it does not depend on fixed locations or costly infrastructure investment.

Despite its potential, developing and validating CP under real-world CATS remains difficult due to the lack of suitable datasets. Existing datasets \cite{yu2022dair,yu2023v2x,xu2023v2v4real,zimmer2024tumtraf,ma2024holovic,xiang2024v2x} have established a pioneering foundation, yet they mainly capture ordinary weather, lighting, and traffic conditions, where single-vehicle perception already works well. As a result, current CP research under CATS largely depends on digital simulations \cite{jiang2024weather,li2024v2x} or indoor mockups \cite{sezgin2023safe}.

While a recent V2I dataset \cite{yang2024v2x} includes certain CATS conditions, such examples remain extremely rare and are limited to static infrastructure viewpoints. In contrast, V2V cooperation involves dynamic agents with diverse viewpoints and continuously changing spatial relations. While this flexibility is key for scalable CP, it also brings challenges in calibration, time synchronization, and temporal alignment, especially in CATS. To date, considering its stringent collecting conditions, no dataset captures real-world V2V perception under CATS.

To fill this gap, we introduce CATS-V2V, the first real-world V2V CP dataset covering CATS. It supports tasks on perception and localization with strict time synchronization, spatial-temporal alignment, and cross-view consistency. Data were collected across ten weather and lighting conditions and ten diverse locations (see \cref{fig:scenarios}) using two experimental vehicles equipped with synchronized sensors, including a 10 Hz LiDAR, seven multi-view 30 Hz cameras, and one 125 Hz inertial navigation system (INS).

The dataset correspondingly provides a time-consistent 3D bounding box annotations with global object IDs and static HD maps for constructing 4D BEV scenes (see \cref{fig:combine}). Furthermore, we propose a target-based temporal alignment method alongside with the dataset, which ensures all objects are precisely aligned across modalities.

Our key contributions are summarized as follows:
\begin{itemize}
\item The first real-world V2V cooperative perception dataset covering diverse weather (sunny, cloudy, rainy, snowy) and lighting (clear, direct sunlight, low light, night) conditions, including dynamic work zones.
\item Broad task support with multiple sensor data (LiDAR, camera, INS) and rich annotations (3D boxes, HD maps, traffic information) including object detection, object tracking, trajectory prediction, localization and mapping.
\item Target-based method enabling alignment across hardware time synchronized high-frame-rate multi-view cameras, LiDAR, and INS, achieving highest data quality.
\end{itemize}

We will release the dataset and code for target-based temporal alignment method to foster further research.
\section{Related work}
\label{sec:relawork}

\begin{table*}[htbp]
  \caption{Comparison of CATS-V2V to some existing V2X cooperative perception datasets.}
  \label{tab:comparison}
  \centering
  \begin{tabular}{c|cccccccccc}
    \toprule
    Dataset & S/R & V2X & CATS & Sync & Align & BEV & Clip$\times$Time & LiDAR & Cam & MV$\times$FPS \\
    \midrule
    V2XSet-w\cite{li2024v2x} & \cellcolor{red!20} Sim & \cellcolor{red!20} V2I & \cellcolor{green!20} \checkmark & \cellcolor{red!20} Soft & \cellcolor{red!20} - & \cellcolor{red!20} - & \cellcolor{green!20} 55$\times$25s & \cellcolor{red!20} 2.8K & \cellcolor{red!20} - & \cellcolor{red!20} - \\
    OPV2V-w\cite{li2024v2x} & \cellcolor{red!20} Sim & \cellcolor{green!20} V2V & \cellcolor{green!20} \checkmark & \cellcolor{red!20} Soft & \cellcolor{red!20} - & \cellcolor{red!20} - & \cellcolor{yellow!20} 73$\times$13s & \cellcolor{red!20} 2.7K & \cellcolor{red!20} - & \cellcolor{red!20} - \\
    DeepAccident\cite{wang2024deepaccident} & \cellcolor{red!20} Sim & \cellcolor{green!20} Both & \cellcolor{green!20} \checkmark & \cellcolor{red!20} Soft & \cellcolor{red!20} - & \cellcolor{red!20} - & \cellcolor{yellow!20} 691$\times$8s & \cellcolor{green!20} 285K & \cellcolor{green!20} 1.71M & \cellcolor{yellow!20} 6$\times$10 \\
    SCOPE\cite{gamerdinger2024scope} & \cellcolor{red!20} Sim & \cellcolor{green!20} Both & \cellcolor{green!20} \checkmark & \cellcolor{red!20} Soft & \cellcolor{red!20} - & \cellcolor{yellow!20} $\circ^a$ & \cellcolor{yellow!20} 44$\times$40s & \cellcolor{green!20} 500K & \cellcolor{green!20} 800K & \cellcolor{yellow!20} 5$\times$10 \\
    Adver-City\cite{karvat2024adver} & \cellcolor{red!20} Sim & \cellcolor{green!20} Both & \cellcolor{green!20} \checkmark & \cellcolor{red!20} Soft & \cellcolor{red!20} - & \cellcolor{red!20} - & \cellcolor{green!20} 110$\times$22s & \cellcolor{green!20} 120K & \cellcolor{yellow!20} 480K & \cellcolor{yellow!20} 4$\times$10 \\ 
    \midrule
    DAIR-V2X\cite{yu2022dair} & \cellcolor{green!20} Real & \cellcolor{red!20} V2I & \cellcolor{yellow!20} $\circ^b$ & \cellcolor{red!20} 30 ms & \cellcolor{yellow!20} Stamp & \cellcolor{red!20} - & \cellcolor{green!20} 100$\times$20s & \cellcolor{yellow!20} 39K & \cellcolor{red!20} 39K & \cellcolor{red!20} 1$\times$20$^c$ \\
    V2X-Seq\cite{yu2023v2x} & \cellcolor{green!20} Real & \cellcolor{red!20} V2I & \cellcolor{red!20} - & \cellcolor{red!20} 30 ms & \cellcolor{red!20} - & \cellcolor{green!20} \checkmark & \cellcolor{green!20} 95$\times$16s & \cellcolor{yellow!20} 30K & \cellcolor{red!20} 30K & \cellcolor{red!20} 1$\times$10 \\
    TUMTraf-V2X\cite{zimmer2024tumtraf} & \cellcolor{green!20} Real & \cellcolor{red!20} V2I & \cellcolor{red!20} - & \cellcolor{red!20} 25 ms & \cellcolor{yellow!20} Stamp & \cellcolor{green!20} \checkmark & \cellcolor{red!20} 8$\times$10s & \cellcolor{red!20} 2K & \cellcolor{red!20} 5K & \cellcolor{red!20} 1$\times$15$^c$ \\
    HoloVIC\cite{ma2024holovic} & \cellcolor{green!20} Real & \cellcolor{red!20} V2I & \cellcolor{red!20} - & \cellcolor{red!20} 25 ms$^d$ & \cellcolor{yellow!20} Stamp & \cellcolor{green!20} \checkmark & \cellcolor{red!20} ?$^e$ & \cellcolor{red!20} ?$^e$ & \cellcolor{red!20} ?$^e$ & \cellcolor{yellow!20} 2*25$^c$ \\
    V2X-Radar\cite{yang2024v2x} & \cellcolor{green!20} Real & \cellcolor{red!20} V2I & \cellcolor{green!20} \checkmark & \cellcolor{red!20} 20 ms & \cellcolor{red!20} - & \cellcolor{red!20} - & \cellcolor{yellow!20} 40$\times$20s & \cellcolor{red!20} 16K & \cellcolor{red!20} 32K & \cellcolor{red!20} 1$\times$10 \\
    V2V4Real\cite{xu2023v2v4real} & \cellcolor{green!20} Real & \cellcolor{green!20} V2V & \cellcolor{red!20} - & \cellcolor{red!20} 50 ms & \cellcolor{yellow!20} Stamp & \cellcolor{green!20} \checkmark & \cellcolor{yellow!20} 67$\times$15s & \cellcolor{yellow!20} 20K & \cellcolor{red!20} 40K & \cellcolor{yellow!20} 2$\times$10 \\
    V2X-Real\cite{xiang2024v2x} & \cellcolor{green!20} Real & \cellcolor{green!20} Both & \cellcolor{red!20} - & \cellcolor{red!20} 50 ms$^f$ & \cellcolor{yellow!20} Stamp & \cellcolor{red!20} - & \cellcolor{yellow!20} 68$\times$12s & \cellcolor{yellow!20} 33K & \cellcolor{yellow!20} 171K & \cellcolor{yellow!20} 4$\times$10 \\
    \midrule
    \textbf{CATS-V2V(ours)} & \cellcolor{green!20} Real & \cellcolor{green!20} V2V & \cellcolor{green!20} \checkmark & \cellcolor{green!20} 1 ms & \cellcolor{green!20} Target & \cellcolor{green!20} \checkmark & \cellcolor{green!20} 100$\times$30s & \cellcolor{green!20} 60K & \cellcolor{green!20} 1.26M & \cellcolor{green!20} 7$\times$30 \\
    \bottomrule
  \end{tabular} \\
  \raggedright{\footnotesize S/R = Sim/Real; V2X = V2I(nfrastructure), V2V, or both; BEV = BEV annotations; MV$\times$FPS = Multi-View cameras on each vehicle$\times$Frame-Per-Second. \\ $^a$ It provides a BEV camera's images but not structural representations. \\ $^b$ It provides limited rainy scenarios as CATS. \\ $^c$ The vehicle camera runs on 20 Hz / 15 Hz / 25 Hz, but only 10 Hz data is sampled for publication. \\ $^d$ Only roadside units are claimed to be synchronized under 5 ms. Considering the characteristics of NTP that it employs, delays $>$25 ms can be expected. \\ $^e$ The scale is unknown, as the figures in their paper are inconsistent, and the dataset is not publicly available. \\ $^f$ The LiDARs triggered synchronously with GPS, but not the cameras. Considering their 10 Hz capture frequency, a 50 ms error can be expected.}
\end{table*}

A variety of V2X datasets have been publicly released, yet they vary greatly in realism and coverage of CATS. While simulation-based datasets can easily model adverse conditions, real-world datasets remain dominated by normal weather and lighting. Moreover, most existing real-world V2X datasets still fall short of the quality in multi-modal richness, synchronization, and alignment achieved by well-known single-vehicle perception datasets such as KITTI \cite{geiger2013vision}, Waymo \cite{sun2020scalability}, and nuScenes \cite{caesar2020nuscenes}. Therefore, in this section, we provide a comprehensive review and comparison of existing simulation-based V2X datasets with CATS and real-world V2X datasets, including both V2V and V2I settings, as summarized in \cref{tab:comparison}.

\subsection{Simulated V2X datasets with CATS}
Simulation-based V2X datasets are typically built on high-fidelity simulators such as CARLA \cite{dosovitskiy2017carla} and OpenCDA \cite{xu2021opencda} to capture virtual sensor data.

OPV2V-w and V2XSet-w \cite{li2024v2x} extend the original OPV2V \cite{xu2022opv2v} and V2XSet \cite{xu2022v2x} datasets to simulate CATS, incorporating virtual weather and lighting variations. Several simulation frameworks \cite{hahner2021fog,kilic2025lidar,hahner2022lidar} were leveraged to emulate fog, rain, and snow effects on LiDAR sensors.

SCOPE \cite{gamerdinger2024scope} is a recently introduced large-scale simulation dataset involving 24 connected and autonomous vehicles (CAVs). It supports both realistic LiDAR and multi-view camera models and further reduces the sim-to-real gap by using two Digital Twin maps reconstructed from real-world environments.

DeepAccident \cite{wang2024deepaccident} and Adver-City \cite{karvat2024adver} focus on safety-critical cooperative perception, reconstructing accident-prone scenarios from real-world crash data with CATS.

Simulation datasets inherently provide accurate localization, perfect annotations, and precise time synchronization via software. Moreover, because sensor data are generated instantaneously, there is no need for high-frame-rate cameras and temporal alignment between modalities. However, they differ significantly from real-world conditions, not only in time, sensor information, and annotation errors, but particularly regarding the physical interaction of CATS with sensor hardware, which simulation cannot yet perfectly reproduce.

\begin{figure*}[htbp]
  \centering
  \includegraphics[width = \linewidth]{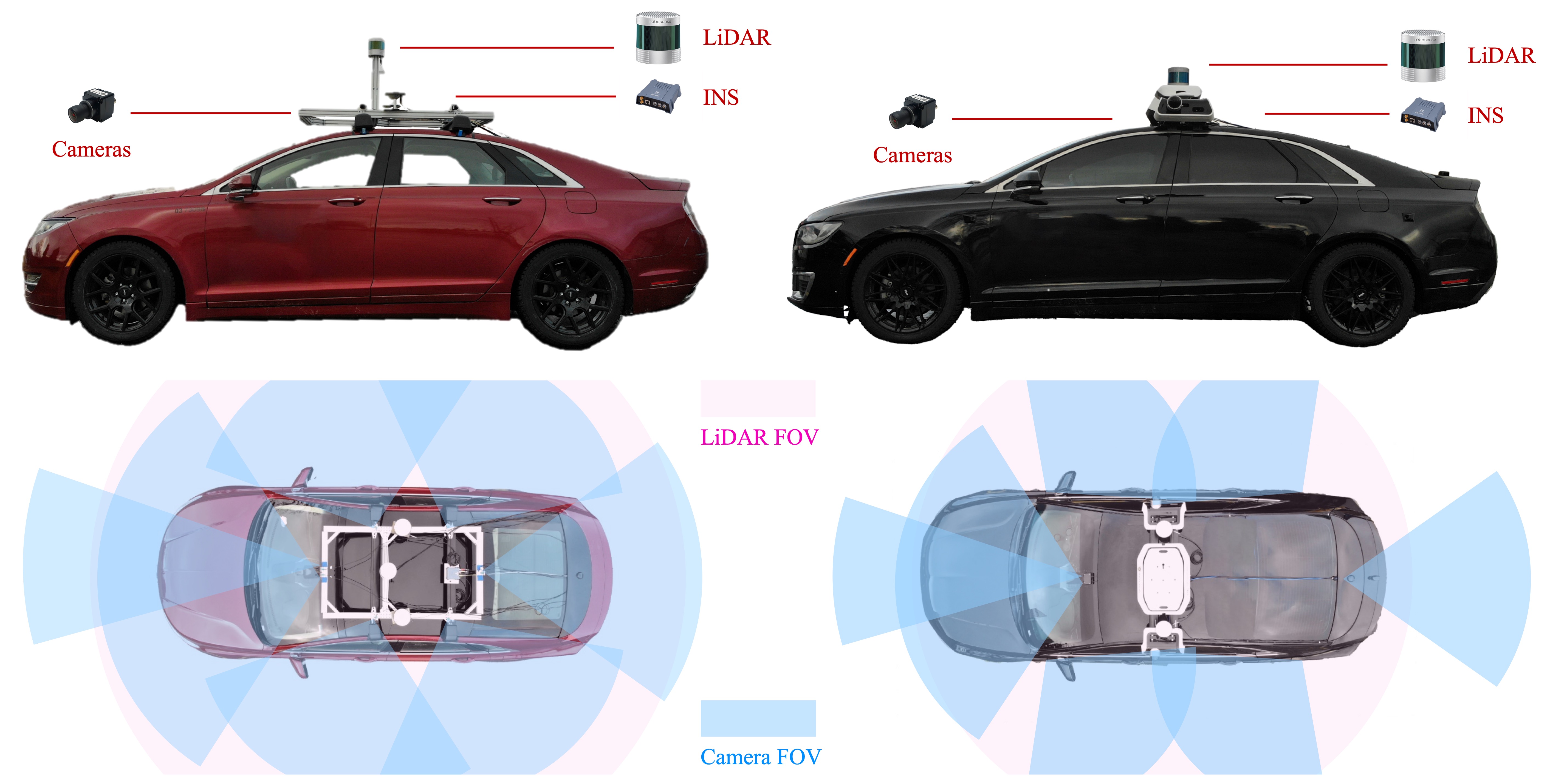}
  \caption{Sensor configurations of our two data-collecting vehicles.}
  \label{fig:vehicle}
\end{figure*}

\subsection{Real-world V2I datasets}
DAIR-V2X \cite{yu2022dair} is one of the first real-world V2X datasets, establishing cooperation between a roadside unit and a vehicle. Subsequently, V2X-Seq \cite{yu2023v2x} extended this setup to consecutive intersections with additional HD map and object tracking annotations. Although the data-collection vehicle in both datasets was equipped with a 20 Hz camera, only compressed front-view images downsampled to 10 Hz based on the timestamps were released. Moreover, both datasets suffer from imperfect time synchronization, with errors exceeding 30 ms.

TUMTraf-V2X \cite{zimmer2024tumtraf} introduces a multi-view intersection dataset consisting of four infrastructure cameras and a single vehicle. The dataset is relatively small, with only eight 10-second clips. Similarly, only the downsampled images from the vehicle’s front-view camera were released. Although annotated for weather and time of day, the dataset only includes two conditions (clear daytime and nighttime) and thus lacks CATS coverage. Their earlier work, TUMTraf-A9 \cite{cress2022tumtrafa9}, captured snow conditions but solely from the infrastructure side, without vehicle cooperation.

HoloVIC \cite{ma2024holovic} is a large-scale dataset covering four intersections equipped with multiple LiDARs and cameras. However, due to limited accessibility and statistical inconsistencies in the paper, information such as the number of clips, segment durations, and data scale remains unclear. It employs Network Time Protocol (NTP) for time synchronization, achieving reported timestamp accuracy of about 5 ms, but only for roadside units. Given the characteristics of NTP, errors exceeding 25 ms are expected between roadside units and the vehicle. Two cameras are also downsampled and aligned by selecting the closest timestamps to the LiDAR frames.

V2X-Radar \cite{yang2024v2x} is the most recent V2I dataset, introducing a radar modality and certain coverage of CATS. However, as a trade-off, it only provides 10 Hz images from one front-view camera, and no alignment could be performed among modalities. It also lacks BEV and corresponding tracking annotations. Although the vehicle platform is described as being hardware-synchronized, timestamp errors of approximately 20 ms were still reported.

\subsection{Real-world V2V datasets}
V2V4Real \cite{xu2023v2v4real} is the first and one of the very few real-world V2V cooperative perception datasets that leverage the perception of two vehicles for collaboration. Each vehicle is equipped with a 32-beam LiDAR and two cameras facing forward and backward, but without hardware-level time synchronization. Images are then aligned to LiDAR frames using the nearest timestamps, which causes some delays. The dataset annotates only five classes of vehicles and does not include vulnerable road users such as pedestrians or cyclists, which may be due to both scene composition and the limited LiDAR resolution.

V2X-Real \cite{xiang2024v2x} is currently the only real-world dataset supporting both V2I and V2V cooperation. It employs 128-beam LiDARs to produce dense point clouds, with LiDARs synchronized to GPS. However, four on-vehicle ZEDi cameras and two Axis infrastructure cameras lack hardware synchronization. As in other V2X datasets, images are aligned to LiDAR frames by closest timestamps. The dataset does not include CATS, or BEV and tracking annotations, and its camera configuration provides limited field-of-view overlap, despite a nominal 360° coverage.
\section{CATS-V2V dataset}
\label{sec:dataset}

Despite these pioneering efforts, existing real-world V2V datasets still lack coverage of CATS. More broadly, current real-world V2X datasets remain limited in quality, with imperfect time synchronization, multi-modal alignment, and annotation richness. In response, we introduce our CATS-V2V dataset to address these gaps. 

The section describes our vehicle configuration, data acquisition, preprocessing, and annotation procedures. More details could be found in the Supplementary Material.

\subsection{Vehicle configurations}
We use two Lincoln MKZ sedans to collect data, as shown in \cref{fig:vehicle}, with sensors integrated through the roof rack. During the data collection process, they are driven by humans.

\subsubsection{Sensors setup}
Each of our two vehicles is equipped with a 128-beam mechanical spinning LiDAR, seven automotive-grade cameras, and one deeply-coupled Inertial Navigation System (INS). The detailed specifications are shown in \cref{tab:sensor}.
\begin{table}[htbp]
  \caption{Detailed sensor specifications on vehicles}
  \label{tab:sensor}
  \centering
  \begin{tabular}{c|p{6.4cm}}
    \toprule
    Sensor & Details \\
    \midrule
    \multirow{6}{*}{LiDAR} & \textbf{Black}: 10 Hz RoboSense Ruby 128-beam LiDAR, dual return mode, 250m range, vertical FOV $-25^\circ \sim 15^\circ$, angle resolution $0.2^\circ$; \\
    & \textbf{Red}: 10 Hz RoboSense Ruby Plus 128-beam LiDAR, dual return mode, 250m range, vertical FOV $-25^\circ \sim 15^\circ$, angle resolution $0.2^\circ$; \\
    \midrule
    \multirow{10}{*}{Camera} & \textbf{Black}: 3 $\times$ (front and rear) 30 Hz OMNIVISION OX08B40, YUV422 8bit, 3840$\times$2160, 140dB HDR, LFM; 4 $\times$ (side) Sony ISX031, YUV422 8bit, 1920$\times$1536, 120dB HDR, LFM;\\
    & \textbf{Red}: 3 $\times$ (front and rear) 30 Hz OMNIVISION OX08B40, YUV422 8bit, 3840$\times$2160, 140dB HDR, LFM; 4 $\times$ (side) OMNIVISION OX03C10, YUV422 8bit, 1920$\times$1080, 140dB HDR, LFM; \\
    \midrule
    \multirow{3}{*}{INS} & Deeply-coupled INS integration with Epson G320 IMU, 1 cm + 1 ppm (RMS) accuracy with RTK, $0.5^\circ$/h bias instability. \\
    \bottomrule
  \end{tabular}
\end{table}

Each vehicle is equipped with a 360° mechanical spinning LiDAR mounted at the top-center position. To mitigate interference from water droplets and snowflakes under CATS, the LiDARs are operated in dual-return mode, providing both the strongest and last points.

Seven cameras are installed on each vehicle, including two front-view (wide-angle and telephoto), one rear-view, and four side-view ones. The specific fields of view (FOVs) are illustrated in \cref{fig:vehicle}. All cameras feature High Dynamic Range (HDR) with LED Flicker Mitigation (LFM) capability, which enhances robustness under CATS.

Each vehicle is also equipped with a high-precision INS connected to a nearby Real-Time Kinematic (RTK) base station within 10 km, leading to an accuracy better than 2 cm (1 cm + 1 ppm). During data collection, both systems maintain a fixed RTK status. In occasionally obstructed environments (e.g., beneath trees or bridges) the tactical-grade IMU ensures short-term localization without drift.

\subsubsection{Time synchronization}
\begin{figure}[htbp]
  \centering
  \includegraphics[width = \linewidth]{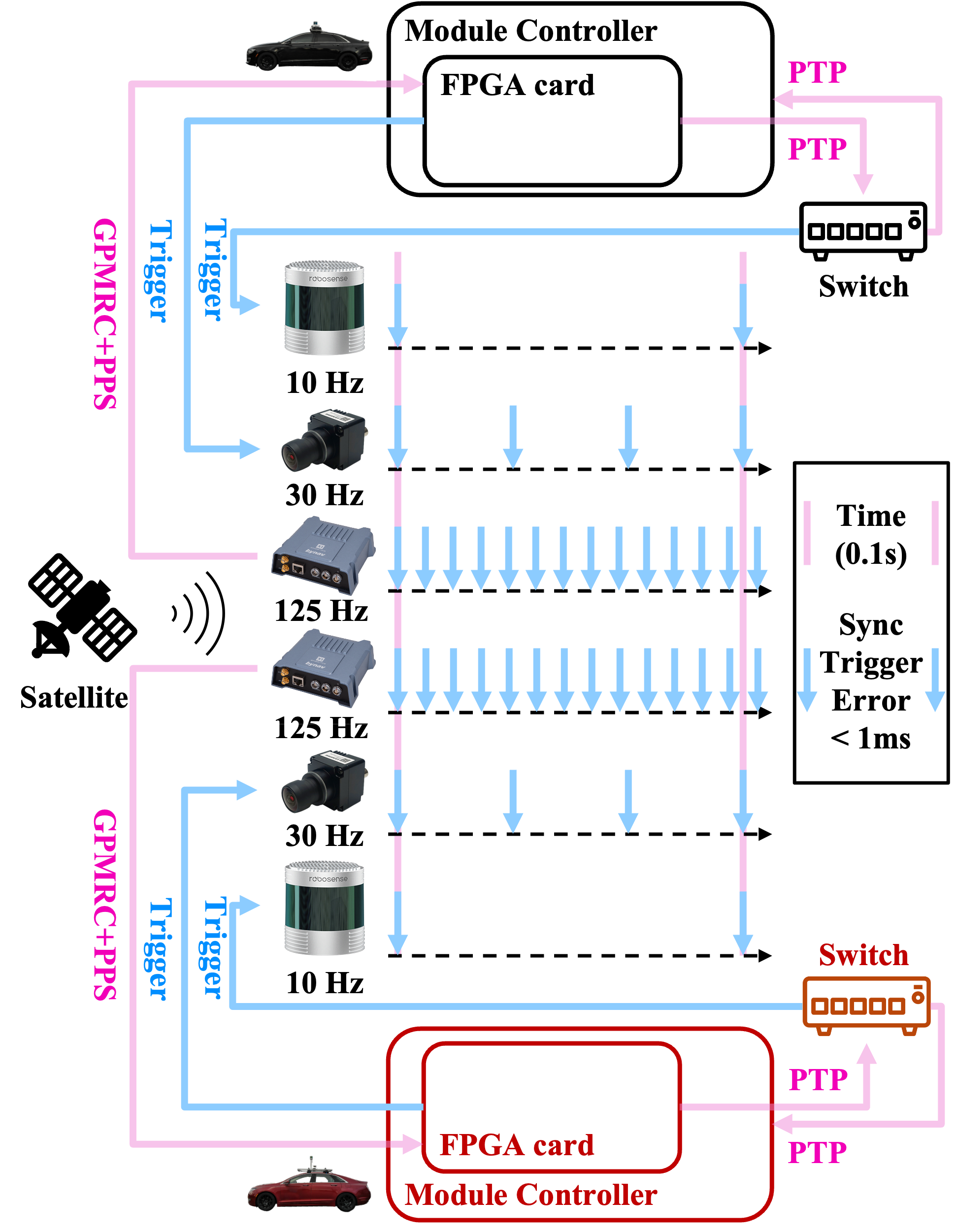}
  \caption{Time synchronization topology for all sensors and module controllers of both vehicles.}
  \label{fig:sync}
\end{figure}

Time synchronization is crucial for multi-modal cooperative perception, as it ensures consistency across all sensors and agents. In our system, all sensors on both vehicles are hardware-synchronized to GPS time and triggered at integer seconds. The overall synchronization topology is illustrated in \cref{fig:sync}.

Each vehicle is equipped with an FPGA card responsible for camera triggering and image acquisition. The card receives the GPRMC (contains absolute time) and PPS (Pulse Per Second, integer trigger) signals from the onboard INS, achieving an offset within 20 ns to the GPS time. It then serves as the master clock, synchronizing all other sensors and module controllers within the local network via the Precision Time Protocol (PTP) through an Ethernet switch, maintaining a time deviation below 1 ms relative to GPS.

Overall, our system achieves a 1 ms synchronization accuracy across all sensors and vehicles, which is an order of magnitude improvement over existing datasets, which typically exhibit over 20 ms offsets, indicating a 40 cm misalignment of objects relatively moving at 72 km/h.

\subsubsection{Calibration}
We use the factory intrinsic calibration for cameras to achieve the highest accuracy, while for extrinsic parameters, we follow two open-source tools \cite{koide2023general,zhu2022robust} and conduct both LiDAR-Camera and LiDAR-INS calibrations.

The LiDAR-Camera extrinsics are estimated by directly aligning the dense LiDAR intensity mapping with the corresponding grayscale camera image \cite{koide2023general}. On the other hand, the LiDAR-INS extrinsics are derived from LiDAR odometry with short-term IMU inertial integration (as the origin of the INS coordinate locates at the IMU) \cite{zhu2022robust}.

\subsection{Data acquisition}
We conduct data collection across ten diverse locations, including highways, arterial roads, collector roads, and local streets (see \cref{fig:scenarios}). For highway environments, we capture both mainline and ramp scenarios. On lower-level roads, our dataset includes signalized, all-way-stop, and unprotected intersections. Residential and campus areas are touched, where vulnerable road users (VRUs) such as pedestrians and cyclists are more frequent.

Each location contains ten repeated runs recorded under varying weather and lighting conditions (see \cref{fig:scenarios}), all referenced to the same HD maps. All sensor data are recorded using ROS2 bag files. The cameras, LiDARs, and INS each publish hardware timestamps through modified drivers.

\subsection{Preprocessing} \label{sec:preprocess}

We then perform motion compensation for LiDAR frames to deskew distortions caused by ego-motion during scanning (0.1 s at 10 Hz, up to 2 m error at 72 km/h). The process is essential for accurate multi-frame registration and object localization \cite{yoon2019mapless}, yet existing datasets \cite{luo2025mixed,xiang2025v2x} acknowledge but do not address the issue, partially due to the absence of per-point timestamps or precise poses.

To ensure consistent annotations between the two vehicles, we register their point clouds into a unified coordinate system. The transformation between the two LiDAR coordinate systems is derived from the vehicle poses and the LiDAR-INS extrinsics, as expressed in \eqref{eq:register}:
\begin{subequations}
\begin{flalign}
    & T_{L_1 L_2}^{\text{init}} = (T_{I_1 L_1})^{-1} (T_{W I_1})^{-1} T_{W I_2} T_{I_2 L_2}, \label{eq:register} \\
    & T_{L_1 L_2}^{\text{init}} \xrightarrow{\text{GICP}} T_{L_1 L_2}^{\text{refine}}, \label{eq:refine}
\end{flalign}
\end{subequations}
where $T_{W I}$ denotes the pose of the INS in the world coordinate frame, and $T_{I L}$ represents the extrinsic calibration between the INS and the LiDAR.

\begin{figure}[htbp]
  \centering
  \includegraphics[width = \linewidth]{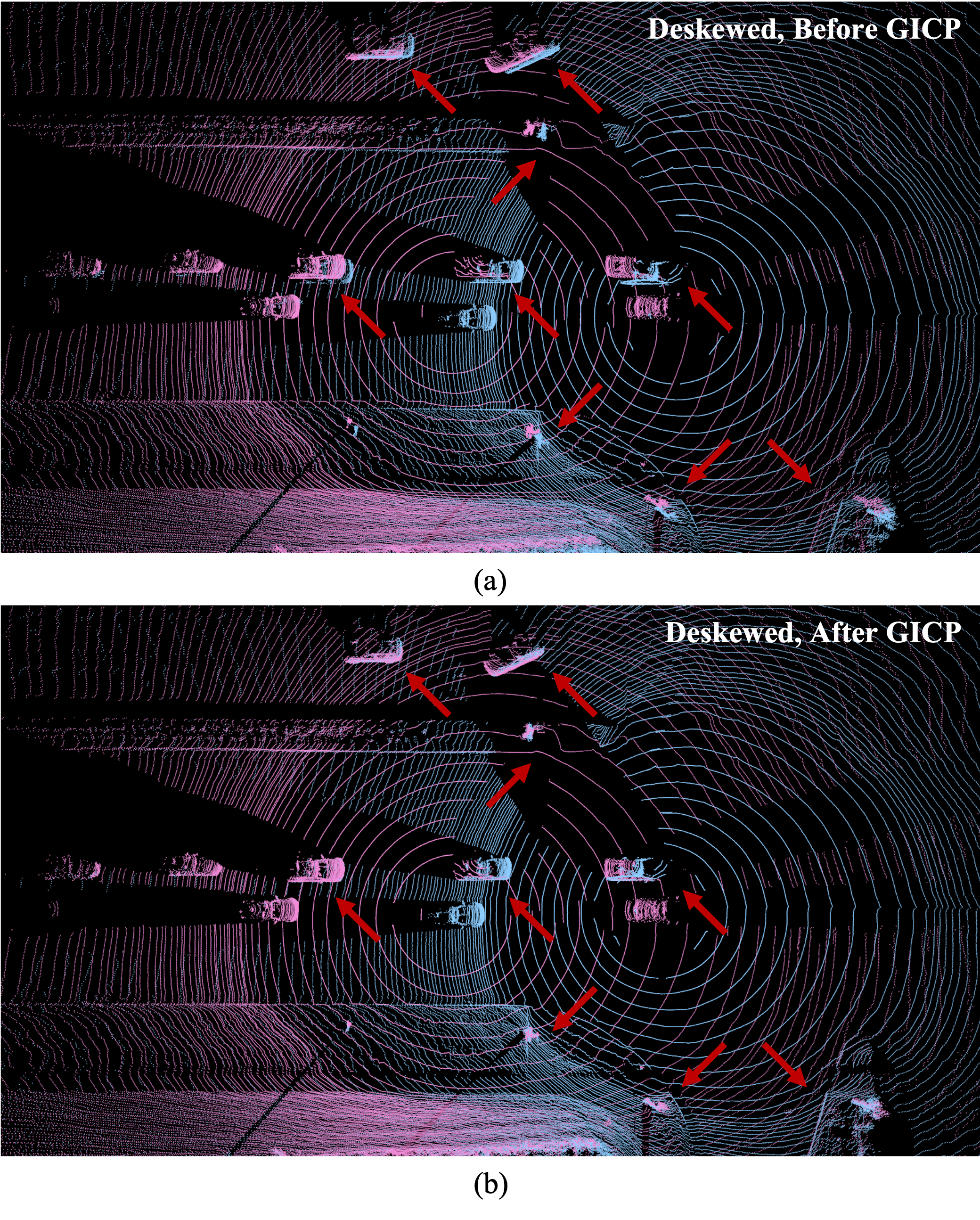}
  \caption{Comparison between two deskewed point clouds register (a) with initial estimation; (b) after GICP refinement.}
  \label{fig:register}
\end{figure}

However, small accumulated errors from four transformations can propagate (see \cref{fig:register}). Therefore, we refine it using Generalized Iterative Closest Point (GICP) \cite{koide2021voxelized} to get $T_{L_1 L_2}^{\text{refine}}$, setting $T_{L_1 L_2}^{\text{init}}$ as the initial estimation.

\subsection{Data annotation}
We provide precise 3D bounding boxes, with time-consistent object dimensions for rigid targets. Each dynamic object is also assigned a globally unique ID, enabling cross-frame and cross-agent re-identification.

Dynamic objects are categorized into Vehicles and VRUs. The Vehicle category includes \textit{Car}, \textit{Van}, \textit{Truck}, \textit{Trailer}, \textit{Bus}, and \textit{Others}; The VRU category includes \textit{Pedestrian}, \textit{Scooter}, \textit{Bicycle}, and \textit{Motorcycle} rider.

Additionally, we introduce a virtual link attribute to indicate physical coupling between two annotated objects. For instance, a link between a \textit{Car} and a \textit{Trailer} object may represent a pickup truck towing a trailer.
\section{Supporting tasks}

\begin{table*}[t]
  \caption{List of supporting tasks of CATS-V2V dataset.}
  \label{tab:task_support}
  \centering
    \begin{tabular}{c|c|>{\centering\arraybackslash}p{3cm}|p{6.5cm}}
    \toprule
    \textbf{Category} & \textbf{Tasks} & \textbf{Modalities} & \textbf{CATs-V2V Support} \\
    \midrule

    \multirow{4}{*}{\textbf{Perception}} 
    & \multirow{2}{*}{2D/3D Detection} 
    & LiDAR; \multirow{1}{*}{Multi-view Camera}  
    & 2D and 3D bounding box annotations \\
    & \multirow{2}{*}{2D/3D Tracking}  
    & LiDAR; \multirow{1}{*}{Multi-view Camera}  
    & 2D and 3D bounding box annotations; Global object IDs across frames\\
    \midrule

    \multirow{5}{*}{\textbf{Spatial Understanding}} 
    & \multirow{2}{*}{Map Generation}
    & LiDAR; IMU/INS; Multi-view Camera
    & High-frequency IMU and RTK-fixed INS; HD maps and BEV annotations\\
    & \multirow{1}{*}{SLAM/Odometry}
    & \multirow{1}{*}{All} 
    & High-frequency IMU and RTK-fixed INS\\
    & \multirow{2}{*}{3D Reconstruction}
    & \multirow{2}{*}{All}
    & HD Maps and geometry transformations based-on RTK-fixed INS \\
    \midrule

    \multirow{2}{*}{\textbf{Multi-Modal Learning}} 
    &  \multirow{2}{*}{Joint Compression} 
    & \multirow{1}{*}{LiDAR;} \multirow{1}{*}{Multi-view Camera} 
    & Hardware-synchronized sensor frames with deskewing and temporal alignment\\
    \midrule

    \multirow{5}{*}{\textbf{Cross-Modal Learning}} 
    & \multirow{3}{*}{Depth Estimation}
    & \multirow{2}{*}{LiDAR;} \multirow{2}{*}{Multi-view Camera} 
    & Geometry transformations based-on RTK-fixed INS; Multi-frame fused point clouds for depth supervision \\
    & View Synthesis
    & IMU/INS; \multirow{1}{*}{Multi-view Camera}
    & Overlapping camera frames; Multi-frame fused point clouds \\
    \midrule

    \multirow{2}{*}{\textbf{Domain Adaptation}} 
    & \multirow{2}{*}{Scene Transfer} 
    & \multirow{2}{*}{Multi-view Camera} 
    & Filtered sets of frames from identical scenes captured under different conditions \\
    \bottomrule
  \end{tabular}
\end{table*}

Thanks to our hardware-synchronized multi-sensor setup, high-quality temporal alignment, and scale of diverse multi-pass data (detailed in \cref{sec:dataset}), the CATS-V2V dataset is well-suited for a wide range of computer vision and multi-modal perception tasks, especially for rare cases.

In addition to providing high-quality 2D and 3D annotations, we offer a set of task-specific data conversion and segment tools to facilitate research and practical usage across different domains.

In the following \cref{tab:task_support}, we summarize the five major categories of vision tasks\cite{yang2023bevheight++}\cite{10095556}\cite{10757429}\cite{li2024light} supported by CATS-V2V. Detailed data processing pipelines and annotation formats for each task are provided in the Supplementary Material.

Due to the extensive range of supported tasks, we do not provide benchmarking results for every task within this dataset release. Instead, we will introduce comprehensive evaluations in a separate future study based on CATS-V2V.
\section{Target-based temporal alignment} \label{sec:exp}

Despite fine calibrations and millisecond time synchronization, precise temporal alignment across modalities is not guaranteed. In this section, we analyze this limitation and progressively introduce our frame-based and target-based temporal alignment method.

\subsection{Stamp-based alignment}
While only a few real-world V2X datasets are equipped with multi-view or high-frame-rate cameras \cite{yu2022dair,zimmer2024tumtraf,ma2024holovic,xu2023v2v4real,xiang2024v2x}, they mostly report that LiDAR and camera frames are aligned with the closest timestamps (see \cref{tab:comparison}).

\begin{figure*}[htbp]
  \centering
  \includegraphics[width = \linewidth]{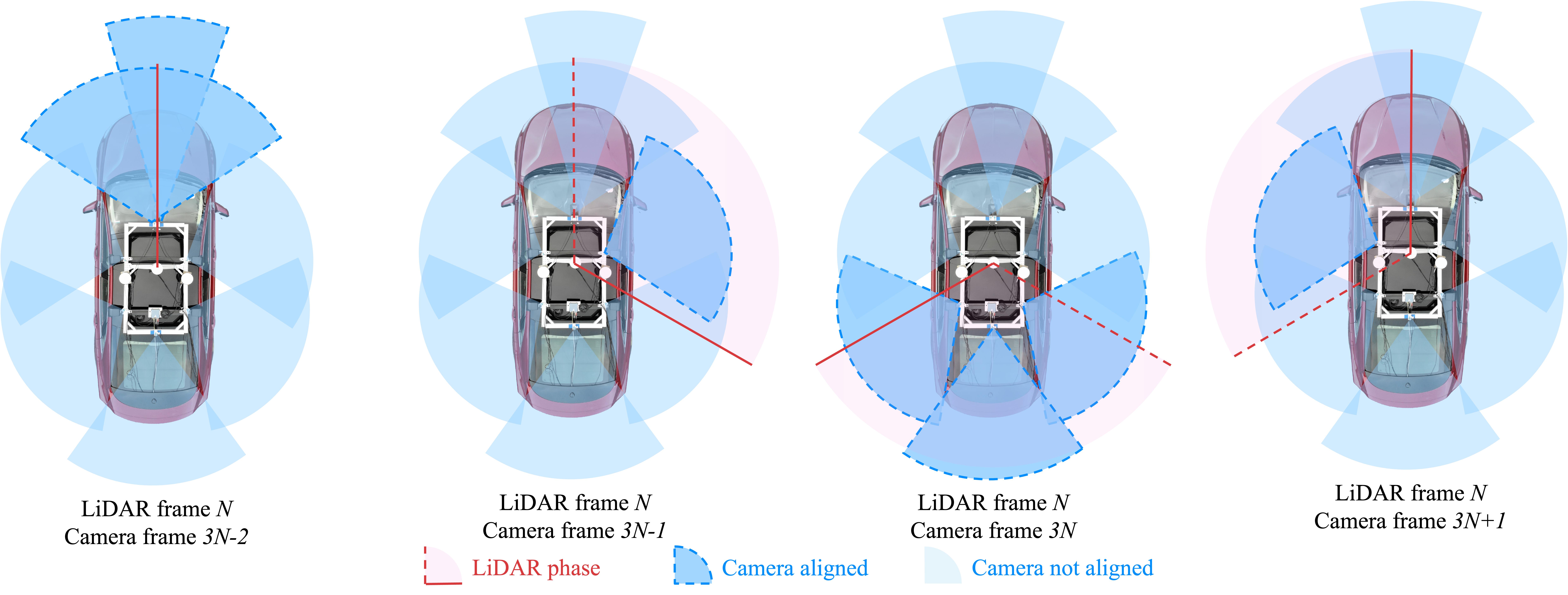}
  \caption{Illustration of alignment between camera and LiDAR frames due to the inherent characteristics of mechanical spinning LiDARs.}
  \label{fig:frame}
\end{figure*}

While this approach provides a convenient approximation, it neglects the inherent characteristics of mechanical spinning LiDARs, which acquire points continuously over a full revolution rather than instantaneously. As a result, each azimuth within a single scan corresponds to a slightly different timestamp. Consequently, stamp-based alignment introduces temporal misalignment, as shown in \cref{fig:frame}.

\subsection{Frame-based alignment}
In contrast, our dataset adopts a frame-based alignment strategy that combines 30 Hz multi-view cameras with LiDAR scans containing per-point timestamps. For each camera, we select the subset of LiDAR points falling within its FOV and associate them with the camera frame whose timestamp is closest to the acquisition times of those points.

We take one of our vehicle's sensor configurations as an example (see \cref{fig:frame}). Since cameras operate at 3 $\times$ the LiDAR frame rate, the two forward cameras are aligned with the starting timestamp of each LiDAR scan, the front-right camera is aligned with the subsequent image frame, the three rear-facing cameras use the next frame, and the front-left camera corresponds to the ending timestamp.

At preprocessing (\cref{sec:preprocess}), we perform motion compensation to correct distortions by compensating all points to the timestamp of the first scan point. This procedure effectively deskews the static background of each LiDAR frame. 

However, for frame-based alignment with cameras, the LiDAR points must further be compensated to the corresponding camera frame timestamp, so that both the static environment and dynamic objects are geometrically consistent across modalities.

\begin{table*}[htbp]
  \caption{Quantitative evaluation of our proposed temporal alignment methods.}
  \label{tab:eval}
  \centering
  \begin{tabular}{c|ccccc}
    \toprule
    Method base & average IoU & Recall@IoU=0.3 & Recall@IoU=0.5 & Recall@IoU=0.7 & center-offset (px) \\
    \midrule
    Stamp & \cellcolor{red!20} 0.3736 & \cellcolor{red!20} 0.6206 & \cellcolor{red!20} 0.3906 & \cellcolor{red!20} 0.1172 & \cellcolor{red!20} 61.54 \\
    \midrule
    \textbf{Frame (ours)} & \cellcolor{green!10} 0.4493 $(\uparrow20.3\%)$ & \cellcolor{green!10} 0.6795 $(\uparrow9.5\%)$ & \cellcolor{green!10} 0.5766 $(\uparrow47.6\%)$ & \cellcolor{green!10} 0.2494 $(\uparrow113\%)$ & \cellcolor{green!10} 50.26 $(\downarrow18.3\%)$ \\
    \midrule
    \textbf{Target (ours)} & \cellcolor{green!20} \textbf{0.4623} $(\uparrow23.7\%)$ & \cellcolor{green!20} \textbf{0.6932} $(\uparrow11.7\%)$ & \cellcolor{green!20} \textbf{0.5947} $(\uparrow52.3\%)$ & \cellcolor{green!20} \textbf{0.2768} $(\uparrow136\%)$ & \cellcolor{green!20} \textbf{49.76} $(\downarrow19.1\%)$ \\
    \bottomrule
  \end{tabular}
\end{table*}

\subsection{Target-based alignment}
While frame-based alignment greatly improves temporal consistency across modalities, it still suffers from rare misalignment in multi-camera overlap or wide-FOV scenarios.

In practice, a single object may appear simultaneously in multiple camera views whose frame-based alignments correspond to different LiDAR scanning times. Similarly, wide-angle camera images may contain a few objects whose alignment should correspond to a previous or a next frame.

To address this issue, we introduce a target-based temporal alignment strategy. After annotation (or LiDAR-based object detection), we compute the average timestamp of all points belonging to each object and associate the object with the nearest camera frame. The corresponding LiDAR points are then motion-compensated to that timestamp.

\subsection{Quantitative Evaluation}
To quantitatively evaluate our proposed alignment strategies, we conduct thorough experiments on one representative clip selected from the 100 recorded scenes.

This clip is captured under favorable conditions without strong sunlight, or inclement weather, so as to minimize perception noise and annotation uncertainty. We manually annotate 2D bounding boxes for all visible dynamic objects across five camera views (not on the two right-view cameras, as we drive on the right lane and they contain few objects), providing ground truth for alignment evaluation.

For benchmarking, we project the annotated 3D bounding boxes onto the images using three different temporal alignment methods: (1) Stamp-based alignment, (2) Frame-based alignment (ours), and (3) Target-based alignment (ours). The projected boxes are then compared against the manually annotated 2D boxes on the corresponding images.

We evaluate the alignment accuracy using three standard metrics: average IoU, Recall at different IoU levels, and mean center-point deviation between the projected and annotated 2D bounding boxes. As shown in \cref{tab:eval}, higher average IoU and Recall among different IoU levels, along with lower center offset, indicate better temporal alignment across modalities progressively with our proposed frame-based and target-based temporal alignment methods.

\section{Conclusion}
\label{sec:conclusion}
In this work, we present \textbf{CATS-V2V}, the first real-world cooperative perception dataset with \textbf{C}omplex \textbf{A}dverse \textbf{T}raffic \textbf{S}cenarios under the \textbf{V}ehicle-to-\textbf{V}ehicle (V2V) collaboration domain. It is a large-scale dataset collected with two vehicles across ten diverse weather, lighting, and traffic conditions (rain, snow, direct sunlight, nighttime, etc) at ten locations, including highways, arterial roads, and urban intersections in campus and residential areas. Specifically, the dataset provides anonymized point clouds, images, and pose records from one LiDAR, seven cameras, and an INS for each vehicle, as well as time-consistent global-identified 3D bounding box annotations and HD Maps. We additionally propose a target-based temporal alignment method to improve the quality of our dataset along with hardware time synchronization. Results on one selected clip from the dataset have demonstrated that our method significantly improves alignment performance across modalities. Furthermore, we offer task-specific data conversion and segment tools to facilitate our supporting tasks. We hope this largest-scale, most supportive, and highest-quality V2V dataset with CATS to date could promote research in related communities. Future plans include combining roadside infrastructure and diverse emerging automotive sensors to provide a richer dataset covering various corner cases and developing tools for converting it into motion and trajectory datasets.
{
    \small
    \bibliographystyle{ieeenat_fullname}
    \bibliography{main}
}

\end{document}